\documentclass{article}
\usepackage{ijcai/ijcai22}

\usepackage{times}
\usepackage{soul}
\usepackage{url}
\usepackage[hidelinks]{hyperref}
\usepackage[utf8]{inputenc}
\usepackage[small]{caption}
\usepackage{graphicx}
\usepackage{amsmath}
\usepackage{amsfonts}
\usepackage{amsthm}
\usepackage{booktabs}
\usepackage{multirow}
\usepackage{algorithm}
\usepackage{algorithmic}
\usepackage{xcolor}
\usepackage{color}
\usepackage{pifont}
\newcommand{\txx}{{\tt TXx}}
\newcommand{\tnx}{{\tt TNx}}
\newcommand{\txn}{{\tt TXn}}
\newcommand{\tnn}{{\tt TNn}}
\newcommand{\prcptot}{{\tt prcptot}}
\newcommand{\ronemm}{{\tt r1mm}}
\newcommand{\rtenmm}{{\tt r10mm}}
\newcommand{\rtwentymm}{{\tt r20mm}}

\newcommand{\cmark}{\ding{51}}
\newcommand{\xmark}{\ding{55}}
\newcommand{\diag}{\text{diag}}
\title{COMET Flows:  Towards Generative Modeling of \\Multivariate Extremes and Tail Dependence}

\author{
Andrew McDonald$^1$
\and
Pang-Ning Tan$^1$\And
Lifeng Luo$^2$
\affiliations
$^1$Department of Computer Science and Engineering, Michigan State University\\
$^2$Department of Geography, Michigan State University\\
\emails
\{mcdon499, ptan, lluo\}@msu.edu
}

\begin{document}

\maketitle

\begin{abstract}
Normalizing flows---a popular class of deep generative models---often 
fail to represent extreme phenomena observed in real-world processes. In particular, existing normalizing flow architectures struggle to model multivariate extremes, characterized by heavy-tailed marginal distributions and asymmetric tail dependence among variables. In light of this shortcoming, we propose COMET (\underline{CO}pula \underline{M}ultivariate \underline{E}x\underline{T}reme) Flows,  
which decompose the process of modeling a joint distribution into two parts: (i) modeling its marginal distributions, and (ii) modeling its copula distribution. COMET Flows capture heavy-tailed marginal distributions by combining a parametric tail belief at extreme quantiles of the marginals with an empirical kernel density function at mid-quantiles. In addition, COMET Flows
capture asymmetric tail dependence 
among multivariate extremes by viewing such dependence as inducing a low-dimensional manifold structure in feature space. 
Experimental results on both synthetic and real-world datasets demonstrate the effectiveness of COMET Flows in capturing both heavy-tailed marginals and asymmetric tail dependence compared to other state-of-the-art baseline architectures.
All code is available on GitHub.\footnote{https://github.com/andrewmcdonald27/COMETFlows}
\end{abstract}


\section{Introduction}
Modeling multivariate extremes involves two main challenges at odds with the usual assumptions of independence and Gaussianity in learning theory: (i) the challenge of modeling heavy-tailed marginal distributions \cite{Resnick2007}, and (ii) the challenge of modeling tail dependence within the joint distribution \cite{Joe2014}. 
In this work, we propose COMET (\underline{CO}pula \underline{M}ultivariate \underline{E}x\underline{T}reme) Flows---a novel deep generative modeling architecture combining Extreme Value Theory and Copula Theory with normalizing flows---in an effort to more accurately model multivariate extremes exhibiting tail dependence.

Deep generative models estimate an unknown probability distribution from an input data for the purposes of density estimation and sampling. Popularized by advances in generative adversarial networks \cite{Goodfellow2014}, variational autoencoders \cite{Kingma2014}, autoregressive networks and normalizing flows, deep generative models offer a promising path forward as limits to labeled data constrains progress in supervised learning. Among deep generative models, normalizing flows set the gold standard of tractability, allowing exact sampling and exact computation of log-density by modeling a transformation between data vectors ${\bf x} \sim p({\bf x})$ and latent vectors ${\bf z} \sim p({\bf z})$.
Given sufficient data and architectural complexity, normalizing flows are universal density approximators \cite{Kobyzev2020}. 

\begin{figure}
    \centering
    \includegraphics[width=0.4\textwidth]{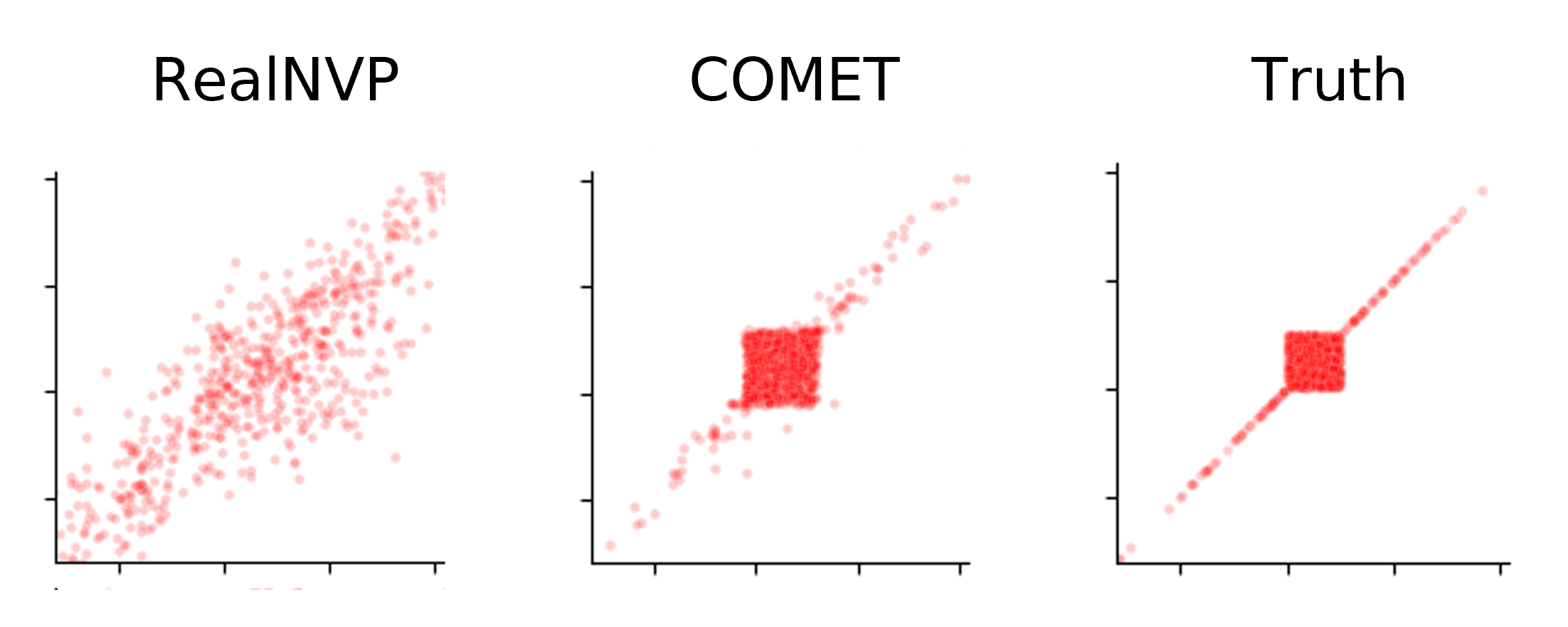}
    \caption{Vanilla normalizing flows such as RealNVP \protect\cite{Dinh2016} struggle to model heavy-tailed, tail-dependent multivariate probability distributions. COMET Flows succeed.}
    \label{fig:motivation}
    \vspace{-0.5em}
\end{figure}

Yet, normalizing flows still struggle to address challenges (i) and (ii) when modeling multivariate extremes (Figure \ref{fig:motivation}). As discussed at length in \cite{Wiese2019,Jaini2020}, vanilla normalizing flows mapping heavy-tailed data distributions 
to light-tailed (e.g., Gaussian) latent distributions 
cannot be Lipschitz-bounded and hence become numerically unstable in practice. Likewise, viewing tail dependence among two or more dimensions 
as inducing a low-dimensional manifold structure in a region of feature space, we may reasonably expect vanilla normalizing flows to perform poorly in that region of space \cite{Cunningham2020,Brehmer2020,Kim2020,Horvat2021}. In this work, we address challenge (i) by employing a copula transformation to decouple 
heavy-tailed marginal distributions $p_i(x_i)$ from the joint dependence structure in $p({\bf x})$, and address challenge (ii) by using a conditional perturbation of inputs during training to capture tail dependence in a stable manner.

\begin{figure*}
    \centering
    \includegraphics[width=0.99\textwidth]{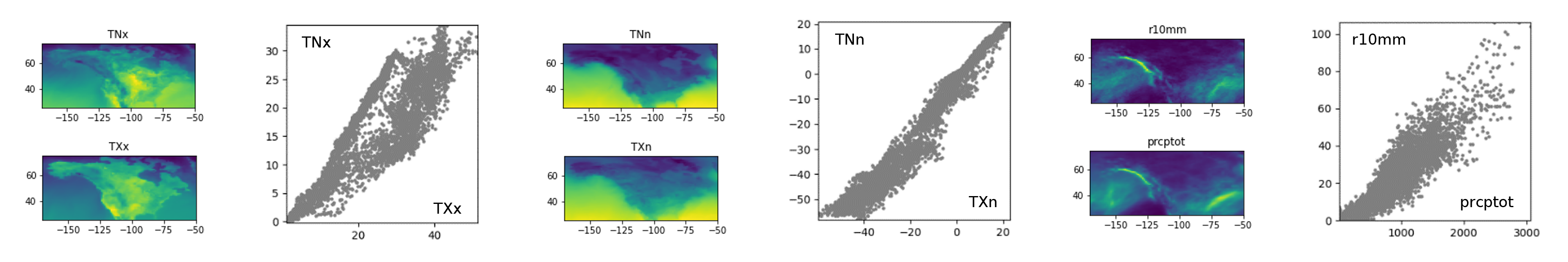}
    \caption{Multivariate extremes in climate science and meteorology---such as \txx, \txn, \prcptot, and \rtenmm in the CLIMDEX \protect\cite{Sillmann2013} dataset---often exhibit heavy-tails and tail dependence that vanilla normalizing flows struggle to model.}
    \label{fig:motivation_climate}
    \vspace{-0.5em}
\end{figure*}

The value of addressing challenges (i) and (ii) is apparent in a number of real-world applications, including climate science and meteorology (Figure \ref{fig:motivation_climate}). Extreme weather has become more frequent and severe in recent years due to climate change. Of particular concern is the projected increase in compound extreme events \cite{Zscheischler2018}, in which simultaneous multivariate extremes pose a greater risk to life and property than comparable univariate extremes spread over time. Accurate deep generative models of such compound extremes allow one to ask how likely an event is to occur, and may even be used to simulate extremes and their effects under various climate change scenarios \cite{Puchko2019,Ayala2020,Klemmer2021}.

Our main contributions are as follows:
\begin{enumerate}
    \item We demonstrate that vanilla normalizing flows parameterized by Lipschitz-bounded neural networks are unable to capture heavy-tailedness and tail dependence in multivariate data;
    \item We present COMET Flows, a novel architecture designed to capture asymmetric heavy-tailedness and tail dependence in multivariate data; and,
    \item We empirically show that COMET Flows present a promising approach for modeling multivariate extremes using synthetic, climate, and benchmark data.

\end{enumerate}

\section{Preliminaries}

\noindent\textbf{Normalizing Flows}
Normalizing flows \cite{Kobyzev2020,Papamakarios2021} offer a systematic 
approach to modeling multivariate probability distributions with deep learning. In particular, normalizing flows facilitate tractable sampling and density estimation under a $d$-dimensional probability distribution $p({\bf x})$ by parameterizing the transformation of each data vector ${\bf x} \in \mathbb{R}^d$ sampled from $p({\bf x})$ to a latent representation ${\bf z} \in \mathbb{R}^d$ following some simple probability distribution $p({\bf z})$ using an invertible neural network $f_{\theta}: \mathbb{R}^d \rightarrow \mathbb{R}^d$. The network is trained to maximize the log-likelihood of observed data $\{ {\bf x}_i \}_{i=1}^N$, which may be computed in closed form using the change-in-variables formula from calculus as a sum of the log-likelihood of the latent vector ${\bf z} = f_{\theta}({\bf x})$ and the log-Jacobian determinant $\log | df_{\theta}({\bf x}) / d{\bf x} |$ of $f_{\theta}$:
\begin{align}\label{eq:change_of_variables}
    p({\bf x}) &= p({\bf z})\left| \frac{df_{\theta}({\bf x})}{d{\bf x}} \right| \\
    \implies  \log p({\bf x}) &= \log p({\bf z}) + \log \left| \frac{df_{\theta}({\bf x})}{d{\bf x}} \right|, \quad \forall {\bf x}, {\bf z} \in \mathbb{R}^d. \nonumber
\end{align}
Following training, the normalizing flow $f_{\theta}$ may be used for density estimation by evaluating $p({\bf x})$ under  \eqref{eq:change_of_variables}. Assuming $f_{\theta}$ is invertible, we may sample from $p({\bf x})$ by sampling ${\bf z} \sim p({\bf z})$ and considering $f_{\theta}^{-1}({\bf z})$.

Architecturally, normalizing flows present two challenges: (i) they must be invertible in order to allow sampling in the generative $({\bf z} \rightarrow {\bf x})$ direction, and (ii) they must allow for efficient computation of the Jacobian determinant $| df({\bf x}) / d{\bf x} |$ in order to be trained and to allow density estimation in the normalizing $({\bf x} \rightarrow {\bf z})$ direction. 
The model we propose in this work satisfies both (i) and (ii) with a coupling-layer based flow backbone \cite{Dinh2016}. 


\vspace{0.5em}\noindent\textbf{Extreme Value Theory}
Extreme Value Theory \cite{Coles2001,Beirlant2005} characterizes the behavior of extremes in precise statistical terms, and may be used to model the realizations of a random variable greater than some threshold. Indeed, under relatively mild assumptions [\citeauthor{Coles2001} \citeyear{Coles2001}, Theorem 4.1] one may show that the excesses of a univariate random variable $x \sim p(x)$ over some threshold $\mu$ follows the generalized Pareto (GP) distribution with probability density function:
\begin{equation}\label{eq:gpd}
    p(x | \mu, \sigma, \xi) = \begin{cases}
    \frac{1}{\sigma}\left[ 1 + \frac{\xi(x-\mu)}{\sigma} \right]^{-1 - 1 / \xi} & \xi \ne 0\\
    \frac{1}{\sigma}\exp \left[ \frac{-(x-\mu)}{\sigma} \right] & \xi=0.\\
    \end{cases}
\end{equation}
Here, $\sigma > 0$ is a scale parameter, $\xi \in \mathbb{R}$ is a tail heaviness parameter, and $p(x)$ has support on $x \in [\mu, \infty)$ for $\xi \geq 0$; alternatively, on $x \in [\mu, \mu - \sigma / \xi]$ for $\xi < 0$. 

Note that the GP distribution may be used to model left excesses (i.e., values under a threshold $\mu$) as right excesses by transforming $x \rightarrow -x$ and $\mu \rightarrow -\mu$. In this work, we utilize the GP distribution to model heavy-tailed marginals of the random vector ${\bf x} \sim p({\bf x})$ in both left and right tails. 


\vspace{0.5em}\noindent\textbf{Copulas}
Copulas \cite{Nelsen2006,Joe2014} describe the dependence structure between uniform random variables, and in particular may be used to model the $d$-dimensional dependence between the components of a multivariate probability distribution after transforming the marginals to be uniformly distributed. Through the lens of Sklar's Theorem [\citeauthor{Nelsen2006} \citeyear{Nelsen2006}, Theorem 2.3.3], a copula $C({\bf u}): [0, 1]^d \rightarrow [0, 1]$ is uniquely defined by a multivariate cumulative distribution function (CDF) $F({\bf x}): \mathbb{R}^d \rightarrow [0, 1]$ having univariate marginal CDFs $F_i(x_i): \mathbb{R} \rightarrow [0, 1]$ through the relationship
\begin{equation}\label{eq:copula_cdf}
    F({\bf x}) = C(F_1(x_1), \ldots, F_d(x_d)), \quad {\bf x} \in \mathbb{R}^d.
\end{equation}
If $F$ is continuous with well-defined marginal inverse CDFs $F_i^{-1}(x_i)$, then \eqref{eq:copula_cdf} may be equivalently expressed as
\begin{equation}\label{eq:copula_cdf_inv}
    C({\bf u}) = F(F_1^{-1}(u_1), \ldots, F^{-1}_d(u_d)), \quad {\bf u} \in \mathbb{R}^d.
\end{equation}
Taking partial derivatives of \eqref{eq:copula_cdf} with respect to each component gives us a similar expression relating $p({\bf x}): \mathbb{R}^d \rightarrow \mathbb{R}$ to the copula density function $c({\bf u}): [0, 1]^d \rightarrow \mathbb{R}$ which is often more convenient in practice when we are interested in density estimation: 
\begin{align}\label{eq:copula_pdf}
    p({\bf x}) &= \frac{\partial^d F({\bf x})}{\partial x_1 \cdots \partial x_d} = \frac{\partial^d C(F_1(x_1), \ldots, F_d(x_d))}{\partial x_1 \cdots \partial x_d} \nonumber \\
    &= c(F_1(x_1), \ldots, F_d(x_d)) \prod_{i=1}^d p_i(x_i), \quad {\bf x} \in \mathbb{R}^d.
\end{align}
In this work, we leverage copulas to model heavy-tailed marginals separately from the dependence structure of the components of a multivariate probability distribution.
    
\vspace{0.5em}\noindent\textbf{Tail Dependence}
Tail dependence \cite{Joe2014} characterizes the degree to which the behavior of two components of a multivariate probability distribution $p({\bf x})$ coincide at high ($u \rightarrow 1^-$) and low ($u \rightarrow 0^+$) quantiles. More formally, let $p({\bf x})$ be a multivariate probability distribution with joint CDF $F$ and marginal CDFs $F_i,\; i=1,\ldots, d$. Then the coefficients of upper and lower tail dependence $\lambda_{ij}^U, \lambda_{ij}^L$ describing the tail relationship of components $i$ and $j$ are given by 
\begin{align}
    \lambda_{ij}^U = \lim_{u \rightarrow 1^-} \mathbb{P}(x_i > F_i^{-1}(u) \; | \; x_j > F_j^{-1}(u)) \label{eq:utd}\\
    \lambda_{ij}^L = \lim_{u \rightarrow 0^+} \mathbb{P}(x_i < F_i^{-1}(u) \; | \; x_j < F_j^{-1}(u)). \label{eq:ltd}
\end{align}
As conditional probabilities, both $\lambda_{ij}^U, \lambda_{ij}^L$ take values in $[0, 1]$. If $\lambda_{ij}^U > 0$, we say that components $i$ and $j$ of $p({\bf x})$ exhibit upper tail dependence; if $\lambda_{ij}^L > 0$, we say that components $i$ and $j$ of $p({\bf x})$ exhibit lower tail dependence. 

In the extreme case of tail dependence between components $x_i$ and $x_j$, it is possible that $F_i^{-1}(u)=x_i=x_j=F_j^{-1}(u)$ for $u < a \approx 0$ or $u > b \approx 1$, such that there exists a local manifold structure of dimension $\mathbb{R}^{d-1}$ embedded in $\mathbb{R}^d$ in the $ij$-th tail of $p({\bf x})$. More generally, if there exists such co-linearity between $k$ components of ${\bf x}$ in the upper or lower tail of $p({\bf x})$, the support of $p({\bf x})$ may be a manifold of dimension $\mathbb{R}^{d-k}$ with measure zero in $\mathbb{R}^d$. Such manifolds are difficult to model using normalizing flows given the requirement that the learned mapping $f_{\theta}: \mathbb{R}^d \rightarrow \mathbb{R}^d$ be invertible---in fact, if $f_{\theta}$ successfully maps a manifold of dimension $\mathbb{R}^{d-k} \subset \mathbb{R}^d$ to a latent space of $\mathbb{R}^d$, it necessarily has an exploding Jacobian determinant. 
As a result, the problem of manifold learning with normalizing flows remains an active area of research \cite{Cunningham2020,Brehmer2020,Kim2020,Horvat2021}---yet to the best of the authors' knowledge, this work is the first to consider the connection between manifold learning, normalizing flows, and tail dependencies in multivariate probability distributions.


\section{Related Work}

\noindent\textbf{Extremes and Normalizing Flows}
\cite{Wiese2019,Jaini2020} prove the inability of normalizing flows to capture heavy-tailed marginal distributions, showing that any mapping 
of heavy-tailed distributions 
to light-tailed (e.g., Gaussian) distributions 
cannot be Lipschitz-bounded. To address this issue, \cite{Wiese2019} propose \emph{Copula \& Marginal Flows} (CM Flows) which disentangle heavy-tailed marginals from the joint distribution by employing a hybrid deep-parametric \emph{marginal flow} mapping inputs $x_i \in \mathbb{R}$ to uniform space $u_i \in [0, 1]$, followed by a RealNVP-inspired \cite{Dinh2016} \emph{copula flow}, mapping the (non-uniform) copula distribution 
to a latent uniform distribution. 
However, the authors omit the implementation of marginal flows in experiments, consider only synthetic bivariate data on $[0, 1]^d$, and suggest the use of \emph{Deep Dense Sigmoidal Flows} \cite{Huang2018} for constructing marginal flows despite the fact these are not analytically invertible and hence do not permit easily tractable sampling. 

\begin{figure*}[t!]
    \centering
    \includegraphics[width=\textwidth]{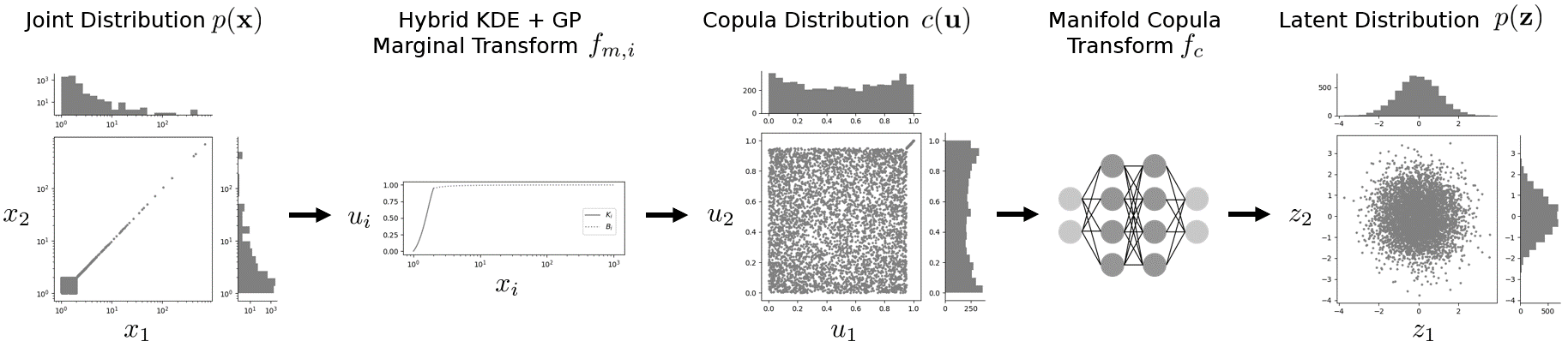}
    \caption{Architecture of COMET Flows, a normalizing flow architecture designed to capture asymmetric and heavy-tailed marginals in $f_m$, along with asymmetric and manifold-like tail dependence structures in $f_c$. The composition $f_c \circ f_m$ allows for tractable log-density estimation, while the composition of analytical inverses $f_m^{-1} \circ f_c^{-1}$ allows for tractable sampling.}
    \label{fig:COMET}
    \vspace{-0.5em}
\end{figure*}

Alternatively, \cite{Jaini2020} propose \emph{Tail Adaptive Flows} (TAFs) which replace the latent isotropic Gaussian $p({\bf z})$ with a latent isotropic Student's $t$ distribution having learnable degrees-of-freedom parameter $\nu>1$, arguing that mapping heavy tails of the input distribution $p({\bf x})$ to heavy tails of the latent distribution will maintain Lipschitz-boundedness. However, the degrees of freedom parameter $\nu$ and the log-likelihood of the data under the model (which depend on one another) must be optimized simultaneously. 
Furthermore, 
by using an isotropic Student's $t$ distribution in latent space, Tail Adaptive Flows implicitly assume symmetry in tail heaviness and hence are unable to capture asymmetric tail heaviness across dimensions. 

\vspace{0.5em}\noindent\textbf{Copulas and Normalizing Flows}
\cite{Bossemeyer2020} develops a generalization of CM Flows to higher dimensions using vine copulas \cite{Czado2019}. 
Orthogonally, \cite{Kamthe2021} present \emph{Copula Flows} motivated by the task of synthetic data generation, and utilize Neural Spine Flows \cite{Durkan2019} to design a generative model able to handle both discrete and continuous data. In \cite{Laszkiewicz2021}, the authors propose to optimize a normalizing flow against a copula distribution in latent space instead of a standard isotropic Gaussian. Other 
relevant works at the intersection of generative modeling and copulas include \cite{Ling2020}, which considers deep estimation of Archimedean copulas, and \cite{Tagasovska2019}, in which vine copulas are fit upon the latent space of an autoencoder to construct a generative model.

\vspace{0.5em}\noindent\textbf{Manifolds and Normalizing Flows}
A novel contribution of our work is to view tail dependence in multivariate probability distributions as inducing a local manifold structure in feature space. Recent works on manifold learning in 
normalizing flows entails density estimation on a $d$-dimensional manifold embedded in $D > d$ dimensional space. 
\emph{Noisy Injective Flows} presented in \cite{Cunningham2020} utilize stochastic inversion for sampling. \emph{M-Flows} are developed in \cite{Brehmer2020} to carry out simultaneous manifold learning and density estimation. \emph{SoftFlow} 
\cite{Kim2020} estimates a conditional distribution of data given a perturbation noise level as input to the flow. This idea is advanced by \cite{Horvat2021} who derive theoretical conditions under which it is possible to recover the underlying manifold using a noising \emph{inflation-deflation} approach.

\section{Proposed Framework: COMET Flows}
Vanilla normalizing flows struggle to capture (i) the heavy-tailed marginals and (ii) asymmetric tail dependencies in multivariate probability distributions. To address these weaknesses, we propose COMET (\underline{CO}pula \underline{M}ultivariate \underline{E}x\underline{T}reme) Flows, a normalizing flow architecture that decouples marginal distributions 
from the joint dependence structure within a multivariate probability distribution 
using a copula, models heavy-tailed marginals using a combination of kernel density and generalized Pareto (GP) distribution functions, and captures tail dependence by considering cases of $\lambda_{ij}^U > 0,\; \lambda_{ij}^L  > 0$ as inducing a low-dimensional manifold in feature space. 

COMET Flows has two main architectural components: a marginal transform $f_m: \mathbb{R}^d \rightarrow [0, 1]^d$ mapping each marginal distribution $p_i(x_i)$ to a uniform distribution such that the dependence structure among dimensions is expressed as a copula, and a copula transform $f_c: [0, 1]^d \rightarrow \mathbb{R}^d$ mapping this copula distribution 
to a tractable latent distribution. 
Thus, COMET Flows define the normalizing flow $({\bf x} \rightarrow {\bf z})$ by the composition $f_c \circ f_m$, and the generative flow $({\bf z} \rightarrow {\bf x})$ by the composition $f_m^{-1} \circ f_c^{-1}$. We describe the details of each component in the following subsections, and present a schematic diagram of COMET Flows in Figure \ref{fig:COMET}.

\vspace{0.5em}\noindent\textbf{Marginal Transform $f_m$} 
In order to capture heavy-tailed marginal distributions $p_i(x_i)$, we construct the marginal transform $f_m$ as a semi-parametric mixture in which the center of each marginal distribution is modeled using a 1-dimensional Gaussian kernel density function, and the tails are modeled using a parametrically-fitted GP distribution motivated by the threshold exceedance paradigm of EVT.

More formally, let $a_i, b_i \in [0, 1]$ be the lower and upper quantiles beyond which to apply parametric GP tails in dimension $i$, and let $\alpha_i, \beta_i \in \mathbb{R}$ be the values in data space corresponding to $a_i, b_i$ obtained through an inverse empirical CDF transform. Then $f_m: \mathbb{R}^d \rightarrow [0, 1]^d$ is defined to be a decoupled mapping acting on each dimension $i=1,\ldots, d$ of its input independently of other dimensions $j \ne i$, with $i$-th component $f_{m, i}$ given by
\begin{equation}\label{eq:marginal_transform}
    f_{m, i}(x_i) = \begin{cases}
        A_i(x_i) & x_i < \alpha_i \\
        K_i(x_i) & x_i \in [\alpha_i, \beta_i] \\
        B_i(x_i) & x_i > \beta_i.
    \end{cases}
\end{equation}
Here, $A_i$ and $B_i$ are the rescaled CDFs of parametric GP distributions fit to the data in the left and right tails of $p_i(x_i)$, respectively, and $K_i$ is the rescaled CDF of a 1D kernel density estimate fit to points in the center of $p_i(x_i)$:
\begin{align}
    A_i(x_i) &= a_i (1 - \tilde{A}_i(\alpha_i - x_i)) \\
    K_i(x_i) &= a_i + (b_i - a_i) \left( \frac{\tilde{K}_i(x_i) - \tilde{K}_i(\alpha_i)}{\tilde{K}_i(\beta_i) - \tilde{K}_i(\alpha_i)} \right)\\
    B_i(x_i) &= b_i + (1 - b_i) \tilde{B}_i(x_i - \beta_i).
\end{align}
Here, $\tilde{A}_i$, $\tilde{K}_i$, and $\tilde{B}_i$ refer to the original CDFs fit to points  $x_i < \alpha_i$, $x_i \in [\alpha_i, \beta_i]$, and $x_i > \beta_i$, respectively. By construction, the inverse marginal transform $f_{m,i}^{-1}: [0, 1] \rightarrow \mathbb{R}$ is well-defined by the rescaled inverse CDFs $(A_i^{-1}, \; K_i^{-1}, \; B_i^{-1})$ on $(u_i < a_i, \; u_i \in [a_i, b_i], \; u_i > b_i)$.

Note that the only learnable portion of $f_{m, i}$ are the parameters of the GP distributions used to model each tail, which can be fit using maximum likelihood estimation (e.g., using \texttt{scipy.stats.genpareto.fit}). This is by design, and vastly simplifies the learning dynamics of the end-to-end $f_c \circ f_m$ model: the GP parameters of each $f_{m, i}$ may be fit and fixed before $f_c$ is trained. This comes in contrast to the approaches of \cite{Wiese2019,Bossemeyer2020}, who allow each $f_{m, i}$ to be learnable flows but struggle to successfully train the coupled $f_c \circ f_m$ model.

\vspace{0.5em}\noindent\textbf{Copula Transform $f_c$}
Given a copula distribution in the unit hypercube $[0, 1]^d$, we construct $f_c$ to be a normalizing flow with codomain $\mathbb{R}^d$. To guarantee that the generative direction $({\bf z} \rightarrow {\bf u})$ maps vectors into the unit hypercube, we apply a logit transform as the first layer of $f_c$ such that the last layer in the generative direction comprises a sigmoid transform. As the logit transform has no learnable parameters, we use $f_c$ to refer only to the learnable transformation applied after the logit transform in what follows.

The learnable transformation $f_c$ is implemented as a coupling layer-based RealNVP \cite{Dinh2016}, repeatedly alternating dimensions $1$:$k$ and $k+1$:$d$ and applying
\begin{align}
    {\bf y}_{1:k} &= {\bf x}_{1:k}\\
    {\bf y}_{k+1:d} &= {\bf x}_{k+1:d} \odot \exp(s({\bf x}_{1:d})) + t(x_{1:d})
\end{align}
in each layer, where $s, t: \mathbb{R}^d \rightarrow \mathbb{R}^k$, $1 < k < d$, and $\odot$ denotes elementwise multiplication. With ${\bf I}_k$ representing the $k \times k$ identity matrix and $*$ being any arbitrary $(d-k) \times k$ matrix, the Jacobian of each coupling layer takes the form
\begin{align}
    {\bf J} = \begin{bmatrix}
        {\bf I}_k & {\bf 0}\\
        * & \diag(\exp[s({\bf x}_{1:d})])
    \end{bmatrix}
\end{align}
and has a readily-computable log determinant $\sum_{i=1}^d s({\bf x}_{1:d})_i$. Because this layer is invertible with tractable log-Jacobian determinant regardless of the form of functions $s$ and $t$, we implement $s$ and $t$ as concatsquash \cite{Onken2020} networks conditioned on a contextual input $\sigma$:
\begin{align}
    s({\bf x}_{1:d}, \sigma) &= s_w(\sigma) s_x({\bf x}_{1:d}) + s_b(\sigma)\\
    t({\bf x}_{1:d}, \sigma) &= t_w(\sigma) t_x({\bf x}_{1:d}) + t_b(\sigma).
\end{align}

To justify this construction, we recall that modeling multivariate extremes presents two main challenges: capturing (i) heavy-tailed marginal distribution and (ii) asymmetric tail dependencies. The marginal transform $f_m$ presented previously addresses (i), leaving the copula transform $f_c$ to address (ii). We accomplish this by viewing tail dependencies as inducing a manifold structure within subspaces of the unit hypercube, and draw inspiration from SoftFlow \cite{Kim2020}.

More precisely, we interpret the conditioning input $\sigma$ to the concatsquash networks $s, t$ to be a noise level, and perturb inputs ${\bf x}$ to $s$ and $t$ by ${\bf x} + {\bf \epsilon}$ where ${\bf \epsilon} \sim \mathcal{N}_d({\bf 0}, \sigma I_d)$ for a noise level chosen uniformly at random $\sigma \sim U[0, \sigma_{\max}]$. This noise level $\sigma$ is passed to the flow as a conditioning contextual input through networks $s$ and $t$ throughout training, allowing the network to learn the conditional distribution $p({\bf x} | \sigma)$ while sidestepping the exploding gradients introduced by the manifold structure of tail dependencies. Yet, this approach still allows the network to recover such tail dependencies at inference time by setting $\sigma = 0$.

\begin{figure}[t!]
    \centering
    \includegraphics[width=0.5\textwidth]{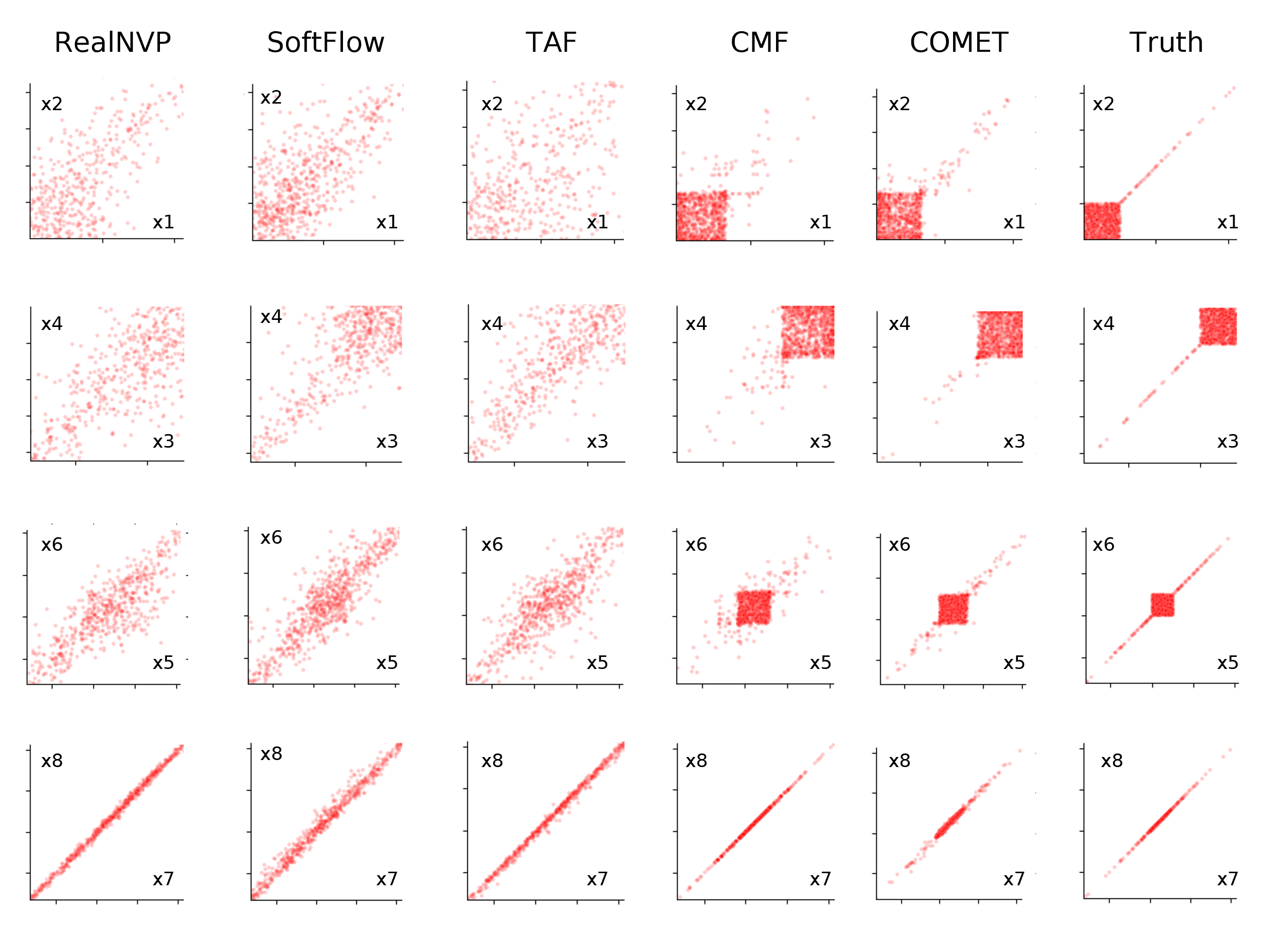}
    \caption{Samples from models trained on the 8D synthetic dataset. Note that RealNVP and SoftFlow blur the manifold structure of tail dependence and exhibit light tails; TAFs generate noisy samples; and CMFs capture structure well but exhibit light tails in comparison to COMET Flows.}
    \label{fig:synthetic}
    \vspace{-0.5em}
\end{figure}

\section{Experimental Evaluation}

\subsection{Baselines}

\vspace{0.5em}\noindent\textbf{RealNVP} A coupling layer-based RealNVP \cite{Dinh2016} implementation optimized against an isotropic Gaussian latent distribution.

\vspace{0.5em}\noindent\textbf{SoftFlow} A modified SoftFlow \cite{Kim2020} implementation in which the FFJORD \cite{Grathwohl2019} backbone is replaced with a RealNVP backbone.

\vspace{0.5em}\noindent\textbf{Tail Adaptive Flow (TAF)} A modified TAF \cite{Jaini2020} implementation composed of a RealNVP flow backbone with an isotropic Student's $t$ latent distribution, parameterized by degrees-of-freedom parameter $\nu$. We consider $\nu$ to be a fixed hyperparameter as no algorithm is presented to learn $\nu$ in the original work.

\vspace{0.5em}\noindent\textbf{Copula Marginal Flow (CMF)} A modified CMF \cite{Wiese2019,Bossemeyer2020} implementation in which the unspecified tail distribution and non-analytically-invertible Deep Dense Sigmoidal Flows \cite{Huang2018} used for $f_m$ are replaced with an analytically-invertible Gaussian KDE and generalized Pareto tail mixture model for $f_m$.

\begin{table*}[ht!]
\centering
\caption{Comparison of baselines and proposed model performance, measured by average negative log-likelhood on a held-out test set. All models are trained until validation loss fails to improve for two consecutive epochs. Standard deviations in NLL over three training runs for each model were all $<\pm 0.1$ and are omitted for space. Columns HT and TD denote whether each model is designed to handle heavy tails and tail dependence, respectively.}
{\small
\begin{tabular}{@{}ccccccccc@{}}
\toprule
\multicolumn{5}{c}{\bf Model} & \multicolumn{4}{c}{\bf NLL}\\ \cmidrule(lr){1-5}\cmidrule(lr){6-9}
{\bf Name} & {\bf Copula} & {\bf Latent} & {\bf HT} & {\bf TD} & {\bf Synthetic} & {\bf CLIMDEX} & {\bf POWER} & {\bf GAS} \\ \midrule
RealNVP & {\color{red} \xmark} & Gaussian & {\color{red} \xmark} & {\color{red} \xmark} & 7.483 & 11.036 & 5.754 & 7.425 \\ \midrule
SoftFlow & {\color{red} \xmark} & Gaussian & {\color{red} \xmark} & {\color{green} \cmark} & 7.563 & 10.765 & 5.790 & 7.484 \\ \midrule
\multirow{3}{*}{TAF} & \multirow{3}{*}{\color{red} \xmark} & Student ($\nu = 16$) & \multirow{3}{*}{\color{green} \cmark} & \multirow{3}{*}{\color{red} \xmark} & 7.046 & 10.679 & 5.500 & 6.790 \\
 &  & Student ($\nu = 32$) &  &  & 7.169 & 10.606 & 5.615 & 7.099 \\
 &  & Student ($\nu = 64$) &  &  & 7.483 & 10.568 & 5.690 & 7.252 \\ \midrule
\multirow{3}{*}{CMF} & (0.01, 0.99) & \multirow{3}{*}{Gaussian} & \multirow{3}{*}{\color{green} \cmark} & \multirow{3}{*}{\color{red} \xmark} & -11.500 & -5.847 & -11.492 & -16.349 \\
 & (0.05, 0.95) &  &  &  & -9.816 & -3.072 & -10.518 & -15.354 \\
 & (0.10, 0.90) &  &  &  & -8.175 & -1.482 & -9.793 & -13.424 \\ \midrule
\multirow{3}{*}{COMET} & (0.01, 0.99) & \multirow{3}{*}{Gaussian} & \multirow{3}{*}{\color{green} \cmark} & \multirow{3}{*}{\color{green} \cmark} & -9.592 & -6.729 & -10.955 & -15.944 \\
 & (0.05, 0.95) &  &  &  & -7.716 & -3.723 & -10.387 & -15.051 \\
 & (0.10, 0.90) &  &  &  & -6.023 & -2.300 & -9.643 & -13.116 \\ \bottomrule
\end{tabular}}
\label{tab:results}
\vspace{-1em}
\end{table*}

\subsection{Data}

\vspace{0.5em}\noindent\textbf{Synthetic Data} 
We design a synthetic dataset exhibiting heavy tails and asymmetric, heterogeneous tail dependence by sampling from a uniform distribution $U$ on $[0, 1]^8$ with probability 0.95, and from independent univariate GP marginal distributions $X$ parameterized by $(\mu, \sigma, \xi) = (0, 1, 1)$ with probability 0.05. These samples are combined such that $x_1, x_2$ exhibit a heavy upper tail with tail dependence; $x_3, x_4$ exhibit a heavy lower tail with tail dependence; $x_5, x_6$ exhibit heavy upper and lower tails, each with tail dependence; and $x_7, x_8$ are perfectly collinear with heavy upper and lower tails. We generate 200,000 vectors in ${\bf x} \in \mathbb{R}^8$ for training, 25,000 vectors for validation, and 25,000 for a held-out test set. Key bivariate marginals of this distribution are shown in the rightmost column of Figure \ref{fig:synthetic}.

\vspace{0.5em}\noindent\textbf{Climate Data}
To evaluate the practical applicability of COMET Flows, we apply them to a subset of CLIMDEX\footnote{\url{https://www.climdex.org/}} extreme climate indices \cite{Sillmann2013}.
Motivated by the projected increase in extreme temperatures and rainfall under climate change, we extract data from a model run under the RCP 8.5 scenario of continued high greenhouse gas emissions, and aim to model the joint distribution of such extremes over North America. We consider the region encompassing 25$^{\circ}$N--75$^{\circ}$N and 170$^{\circ}$W--50$^{\circ}$W and simulation data from the years 2091-2100, in which the effects of climate change on extremes will be most pronounced. We choose four indices of extreme temperature (\txx, \tnx, \txn, \tnn), along with four indices of extreme precipitation (\prcptot, \ronemm, \rtenmm, and \rtwentymm), then add latitude and longitude to form a 10-dimensional dataset characterizing joint extremes and their spatial distribution. Indices are reported at every 0.94$^{\circ}$ latitude and 1.25$^{\circ}$ longitude, giving a total of 97 $\times$ 53 $=$ 5,141 vectors ${\bf x} \in \mathbb{R}^{10}$ per year. We use data from 2091-2098 for training, 2099 for validation, and 2100 for testing. A subset of this data was presented in Figure \ref{fig:motivation_climate}.

\vspace{0.5em}\noindent\textbf{Benchmark Data}
To contextualize the performance of COMET Flows, we consider the POWER and GAS benchmark datasets originally presented in \cite{Papamakarios2017}. We follow the standard preprocessing steps provided in the MAF GitHub\footnote{\url{https://github.com/gpapamak/maf}} repository. Note that we do not expect our baseline results to match state-of-the-art negative log-likelihood values reported in \cite{Kobyzev2020} due to the simplified and adjusted architectures we implement.

\subsection{Results}

\vspace{0.5em}\noindent\textbf{Synthetic Data}
As shown in Figure \ref{fig:synthetic} and Table \ref{tab:results}, the copula-based CM and COMET models dominate in both the normalizing (${\bf x} \rightarrow {\bf z}$) and generative (${\bf z} \rightarrow {\bf x}$) directions on synthetic data due to their functional form: by construction, they are naturally able to account for the asymetrically heavy-tailed margins of training data where vanilla normalizing flows cannot, and where TAF are forced to assume symmetry in tail heaviness. In particular, COMET's ability to capture the low-dimensional manifold structure in the upper tail region of feature space is highlighted by its qualitative performance in the generative direction (Figure \ref{fig:synthetic}), and its ability to avoid the instabilities and exploding gradients associated with manifold structures in feature space by learning perturbed manifolds conditional on noise is apparent by its quantitative performance in the normalizing direction (Figure \ref{tab:results}). COMET's edge over CMF is most evident in its ability to capture tail dependence, though both significantly outperform TAF, SoftFlow, and RealNVP.

\vspace{0.5em}\noindent\textbf{Climate Data} 
COMET's advantage in the normalizing direction is more evident in this experiment. Key bivariate marginals in Figure \ref{fig:motivation_climate} appear to exhibit local manifold structures (e.g., between tails of {\txn} and {\tnn}), explaining COMET's superior performance. This observation suggests that explicitly accounting for such manifold structures in the joint tails of multivariate probability distributions is profitable in practice, particularly when redundancy across dimensions is likely to exist as an artifact of the latent physical processes generating the observed data.

\vspace{0.5em}\noindent\textbf{Benchmark Data} 
COMET is not as performant as CMF on the benchmark data in the normalizing direction, likely due to the absence of heavy-tails and tail dependence in POWER and GAS. Nevertheless, COMET outperforms TAF, SoftFlow, and RealNVP, suggesting that it remains competitive even on datasets it is not explicitly designed to handle.

\section{Conclusion}
Modeling multivariate extremes involves two fundamental challenges: (i) modeling heavy-tailed marginal distributions, and (ii) modeling tail dependence within the joint distribution. To address these challenges, we propose COMET Flows, a novel normalizing flow architecture which draws inspiration from Extreme Value Theory and Copula Theory, and which views tail dependence as inducing a local manifold structure in feature space. We demonstrate the efficacy of COMET Flows through experiments on synthetic, climate, and benchmark datasets, observing competitive performance in sampling and density estimation.



\section*{Acknowledgements}

This research is partially supported by the National Science Foundation under grant IIS-2006633. Any use of trade, firm, or product names is for descriptive purposes only and does not imply endorsement by the U.S. Government. The authors thank Asadullah Hill Galib and Tyler Wilson for their input in discussions regarding the paper.

\bibliographystyle{ijcai/ijcai22}
\bibliography{2.camera}

\begin{thebibliography}{}

\bibitem[\protect\citeauthoryear{Ayala \bgroup \em et al.\egroup
  }{2020}]{Ayala2020}
Alexis Ayala, Christopher Drazic, Brian Hutchinson, Ben Kravitz, and Claudia
  Tebaldi.
\newblock {Loosely Conditioned Emulation of Global Climate Models With
  Generative Adversarial Networks}.
\newblock {\em arXiv}, 2020.

\bibitem[\protect\citeauthoryear{Beirlant \bgroup \em et al.\egroup
  }{2005}]{Beirlant2005}
Jan Beirlant, Yuri Goegebeur, Jozef Teugels, Johan Segers, Daniel {De Waal},
  and Chris Ferro.
\newblock {\em {Statistics of Extremes: Theory and Applications}}.
\newblock Springer, 2005.

\bibitem[\protect\citeauthoryear{Bossemeyer}{2020}]{Bossemeyer2020}
Leonie Bossemeyer.
\newblock {CM Flows - Copula Density Estimation with Normalizing Flows}.
\newblock 2020.

\bibitem[\protect\citeauthoryear{Brehmer and Cranmer}{2020}]{Brehmer2020}
Johann Brehmer and Kyle Cranmer.
\newblock {Flows for simultaneous manifold learning and density estimation}.
\newblock In {\em NeurIPS}, 2020.

\bibitem[\protect\citeauthoryear{Coles}{2001}]{Coles2001}
Stuart Coles.
\newblock {\em {An Introduction to Statistical Modeling of Extreme Values}}.
\newblock Springer, 2001.

\bibitem[\protect\citeauthoryear{Cunningham \bgroup \em et al.\egroup
  }{2020}]{Cunningham2020}
Edmond Cunningham, Renos Zabounidis, Abhinav Agrawal, Ina Fiterau, and Daniel
  Sheldon.
\newblock {Normalizing Flows Across Dimensions}.
\newblock {\em arXiv}, 2020.

\bibitem[\protect\citeauthoryear{Czado}{2019}]{Czado2019}
Claudia Czado.
\newblock {\em {Analyzing Dependent Data with Vine Copulas}}.
\newblock Springer, 2019.

\bibitem[\protect\citeauthoryear{Dinh \bgroup \em et al.\egroup
  }{2016}]{Dinh2016}
Laurent Dinh, Jascha Sohl-Dickstein, and Samy Bengio.
\newblock {Density estimation using Real NVP}.
\newblock {\em arXiv}, 2016.

\bibitem[\protect\citeauthoryear{Durkan \bgroup \em et al.\egroup
  }{2019}]{Durkan2019}
Conor Durkan, Artur Bekasov, Iain Murray, and George Papamakarios.
\newblock {Neural spline flows}.
\newblock In {\em NeurIPS}, 2019.

\bibitem[\protect\citeauthoryear{Goodfellow \bgroup \em et al.\egroup
  }{2014}]{Goodfellow2014}
Ian Goodfellow, Jean Pouget-Abadie, Mehdi Mirza, Bing Xu, David Warde-Farley,
  Sherjil Ozair, Aaron Courville, and Yoshua Bengio.
\newblock {Generative adversarial networks}.
\newblock {\em arXiv}, 2014.

\bibitem[\protect\citeauthoryear{Grathwohl \bgroup \em et al.\egroup
  }{2019}]{Grathwohl2019}
Will Grathwohl, Ricky~T.Q. Chen, Jesse Bettencourt, Ilya Sutskever, and David
  Duvenaud.
\newblock {FFJORD: Free-form continuous dynamics for scalable reversible
  generative models}.
\newblock In {\em ICLR}, 2019.

\bibitem[\protect\citeauthoryear{Horvat and Pfister}{2021}]{Horvat2021}
Christian Horvat and Jean-Pascal Pfister.
\newblock {Density estimation on low-dimensional manifolds: an
  inflation-deflation approach}.
\newblock {\em arXiv}, 2021.

\bibitem[\protect\citeauthoryear{Huang \bgroup \em et al.\egroup
  }{2018}]{Huang2018}
Chin-Wei Huang, David Krueger, Alexandre Lacoste, and Aaron Courville.
\newblock {Neural autoregressive flows}.
\newblock In {\em ICML}, 2018.

\bibitem[\protect\citeauthoryear{Jaini \bgroup \em et al.\egroup
  }{2020}]{Jaini2020}
Priyank Jaini, Ivan Kobyzev, Yaoliang Yu, and Marcus~A. Brubaker.
\newblock {Tails of lipschitz triangular flows}.
\newblock In {\em ICML}, 2020.

\bibitem[\protect\citeauthoryear{Joe}{2014}]{Joe2014}
Harry Joe.
\newblock {\em {Dependence Modeling with Copulas}}.
\newblock CRC Press, 2014.

\bibitem[\protect\citeauthoryear{Kamthe \bgroup \em et al.\egroup
  }{2021}]{Kamthe2021}
Sanket Kamthe, Samuel Assefa, and Marc Deisenroth.
\newblock {Copula Flows for Synthetic Data Generation}.
\newblock {\em arXiv}, 2021.

\bibitem[\protect\citeauthoryear{Kim \bgroup \em et al.\egroup
  }{2020}]{Kim2020}
Hyeongju Kim, Hyeonseung Lee, Woo~Hyun Kang, Joun~Yeop Lee, and Nam~Soo Kim.
\newblock {SoftFlow: Probabilistic Framework for Normalizing Flow on
  Manifolds}.
\newblock In {\em NeurIPS}, 2020.

\bibitem[\protect\citeauthoryear{Kingma and Welling}{2014}]{Kingma2014}
Diederik~P. Kingma and Max Welling.
\newblock {Auto-encoding variational bayes}.
\newblock {\em arXiv}, 2014.

\bibitem[\protect\citeauthoryear{Klemmer \bgroup \em et al.\egroup
  }{2021}]{Klemmer2021}
Konstantin Klemmer, Sudipan Saha, Matthias Kahl, Tianlin Xu, and Xiao~Xiang
  Zhu.
\newblock Generative modeling of spatio-temporal weather patterns with extreme
  event conditioning.
\newblock In {\em ICLR Workshop on Modeling Oceans and Climate Change}, 2021.

\bibitem[\protect\citeauthoryear{Kobyzev \bgroup \em et al.\egroup
  }{2020}]{Kobyzev2020}
Ivan Kobyzev, Simon Prince, and Marcus Brubaker.
\newblock {Normalizing Flows: An Introduction and Review of Current Methods}.
\newblock {\em {IEEE Transactions on Pattern Analysis and Machine
  Intelligence}}, 2020.

\bibitem[\protect\citeauthoryear{Laszkiewicz \bgroup \em et al.\egroup
  }{2021}]{Laszkiewicz2021}
Mike Laszkiewicz, Johannes Lederer, and Asja Fischer.
\newblock {Copula-Based Normalizing Flows}.
\newblock In {\em ICML Workshop on INNF}, 2021.

\bibitem[\protect\citeauthoryear{Ling \bgroup \em et al.\egroup
  }{2020}]{Ling2020}
Chun~Kai Ling, Fei Fang, and J.~Zico Kolter.
\newblock {Deep Archimedean Copulas}.
\newblock In {\em NeurIPS}, 2020.

\bibitem[\protect\citeauthoryear{Nelsen}{2006}]{Nelsen2006}
Roger~B. Nelsen.
\newblock {\em {An Introduction to Copulas}}.
\newblock Springer, 2006.

\bibitem[\protect\citeauthoryear{Onken and Ruthotto}{2020}]{Onken2020}
Derek Onken and Lars Ruthotto.
\newblock Discretize-optimize vs. optimize-discretize for time-series
  regression and continuous normalizing flows.
\newblock {\em arXiv}, 2020.

\bibitem[\protect\citeauthoryear{Papamakarios \bgroup \em et al.\egroup
  }{2017}]{Papamakarios2017}
George Papamakarios, Theo Pavlakou, and Iain Murray.
\newblock {Masked autoregressive flow for density estimation}.
\newblock In {\em NeurIPS}, 2017.

\bibitem[\protect\citeauthoryear{Papamakarios \bgroup \em et al.\egroup
  }{2021}]{Papamakarios2021}
George Papamakarios, Eric Nalisnick, Danilo~Jimenez Rezende, Shakir Mohamed,
  and Balaji Lakshminarayanan.
\newblock {Normalizing Flows for Probabilistic Modeling and Inference}.
\newblock {\em {Journal of Machine Learning Research}}, 2021.

\bibitem[\protect\citeauthoryear{Puchko \bgroup \em et al.\egroup
  }{2019}]{Puchko2019}
Alexandra Puchko, Robert Link, Brian Hutchinson, Ben Kravitz, and Abigail
  Snyder.
\newblock {DeepClimGAN: A High-Resolution Climate Data Generator}.
\newblock In {\em NeurIPS Workshop on Tackling Climate Change with Machine
  Learning}, 2019.

\bibitem[\protect\citeauthoryear{Resnick}{2007}]{Resnick2007}
Sidney~I. Resnick.
\newblock {\em {Heavy Tail Phenomena: Probabilistic and Statistical Modeling}}.
\newblock Springer, 2007.

\bibitem[\protect\citeauthoryear{Sillmann \bgroup \em et al.\egroup
  }{2013}]{Sillmann2013}
Jana Sillmann, Viatcheslav~V Kharin, FW~Zwiers, Xuebin Zhang, and D~Bronaugh.
\newblock Climate extremes indices in the {CMIP5} multimodel ensemble: Part 2.
  {F}uture climate projections.
\newblock {\em Journal of Geophysical Research: Atmospheres}, 2013.

\bibitem[\protect\citeauthoryear{Tagasovska \bgroup \em et al.\egroup
  }{2019}]{Tagasovska2019}
Natasa Tagasovska, Damien Ackerer, and Thibault Vatter.
\newblock {Copulas as high-dimensional generative models: Vine copula
  autoencoders}.
\newblock In {\em NeurIPS}, 2019.

\bibitem[\protect\citeauthoryear{Wiese \bgroup \em et al.\egroup
  }{2019}]{Wiese2019}
Magnus Wiese, Robert Knobloch, and Ralf Korn.
\newblock {Copula \& Marginal Flows: Disentangling the Marginal from its
  Joint}.
\newblock {\em arXiv}, 2019.

\bibitem[\protect\citeauthoryear{Zscheischler \bgroup \em et al.\egroup
  }{2018}]{Zscheischler2018}
Jakob Zscheischler, Seth Westra, Bart~JJM Van Den~Hurk, Sonia~I Seneviratne,
  Philip~J Ward, Andy Pitman, Amir AghaKouchak, David~N Bresch, Michael
  Leonard, Thomas Wahl, et~al.
\newblock Future climate risk from compound events.
\newblock {\em Nature Climate Change}, 2018.

\end{thebibliography}

\end{document}